\pdfoutput=1
\documentclass[sigconf]{acmart}

\usepackage{hyperref}
\usepackage[hyphenbreaks]{breakurl}

\usepackage{enumitem}
\usepackage{balance} 

\AtBeginDocument{%
  }

\setcopyright{acmlicensed}
\copyrightyear{2023}
\acmYear{2023}

\acmConference[MM '23] {Proceedings of the 31st ACM International Conference on Multimedia}{October 29--November 3, 2023}{Ottawa, ON, Canada.}
\acmBooktitle{Proceedings of the 31st ACM International Conference on Multimedia (MM '23), October 29--November 3, 2023, Ottawa, ON, Canada}
\acmPrice{15.00}
\acmISBN{979-8-4007-0108-5/23/10}
\acmDOI{10.1145/3581783.3612133}

\begin{document}

\title{Co-Salient Object Detection with Semantic-Level Consensus Extraction and Dispersion}


\author{Peiran Xu}
\affiliation{%
  \institution{Peking University}
  \city{Beijing}
  \country{China}}
\email{2301112093@pku.edu.cn}

\author{Yadong Mu}
\authornote{Corresponding author.}
\affiliation{%
  \institution{Peking University}
  \city{Beijing}
  \country{China}}
\email{muyadong@gmail.com}




\begin{abstract}
  Given a group of images, co-salient object detection (CoSOD) aims to highlight the common salient object in each image. There are two factors closely related to the success of this task, namely consensus extraction, and the dispersion of consensus to each image. 
  Most previous works represent the group consensus using local features, while we instead utilize a hierarchical Transformer module for extracting semantic-level consensus. Therefore, it can obtain a more comprehensive representation of the common object category, and exclude interference from other objects that share local similarities with the target object. 
  In addition, we propose a Transformer-based dispersion module that takes into account the variation of the co-salient object in different scenes. It distributes the consensus to the image feature maps in an image-specific way while making full use of interactions within the group. 
  These two modules are integrated with a ViT encoder and an FPN-like decoder to form an end-to-end trainable network, without additional branch and auxiliary loss. The proposed method is evaluated on three commonly used CoSOD datasets and achieves state-of-the-art performance.
\end{abstract}





\maketitle

\section{Introduction}
\label{sec1}
Co-salient object detection (CoSOD) is a group-based image understanding task that aims to capture the common salient object presented in each image within the given group. Due to its wide spectrum of applications in object detection, semantic segmentation, and image retrieval, significant research efforts have been invested within this field over the past decade.

Conventionally, methods for solving this task rely on hand-crafted visual features of the images. 
\cite{CVPR15} is one of the first attempts to apply deep learning to this problem. From then on, a variety of network structures have been proposed and greatly improved the performance of this task. Most of these works could be formalized as a four-stage paradigm, which involves (1) image feature representation through an encoder, (2) consensus extraction performed on a group of image features, (3) the dispersion of consensus to individual image features, and (4) co-saliency mask generation through a decoder. Among them, consensus extraction and dispersion (similar to the term ``summarization and search'' in~\cite{CADC}) are two key steps and have been implemented in diverse ways in previous works. We will give a detailed review on these methods in Section~\ref{sec2}.

In this work, we also follow the four-stage paradigm since it is basically consistent with the human cognitive process. Despite significant breakthroughs achieved in this line of work, we have observed two common issues among the majority of prior research. Firstly, most previous works obtain the consensus by selecting or averaging local features~\cite{GICD, UFO, CoRP, GCoNet, DCFM, TCNet, GCoNet+, MCCL}, which can capture the representative region of the common object. However, it cannot provide complete, semantic-level information on the common object's category. Thus, such methods may fail when encountering large intra-class differences, and may highlight some irrelevant objects that have local similarities to the target object but do not belong to the same category. 
Moreover, the majority of previous methods implement the dispersion of consensus to image features by concatenation, addition, element-wise multiplication, dot-product, or their combination~\cite{IJCAI17, RNN, FCN, AAAI19, GICD, GCN, GCoNet, CoSFormer, DeepACG, DCFM, UFO, TCNet, GCoNet+, CoRP, MCCL}. Nevertheless, the common object may present very different attributes and appearances in different images. Therefore, simply adding or multiplying consensus to the image feature maps in the same way may not reflect this diversity, resulting in poor performance on challenging examples. Though some earlier works~\cite{ICNet, CADC} have taken the image-specific variation of consensus into account explicitly or implicitly, they fail to maintain a compact and consistent consensus representation, which may lead to inconsistent detection targets and increased computation. We argue that it is important to balance the commonality and the specificity, while controlling the computational complexity in a reasonable range.

Based on these considerations, we develop two novel blocks for semantic-level comprehensive consensus extraction and image-specific dispersion. In particular, a hierarchical Transformer module is designed for consensus extraction. It alternates between aggregating each image's salient object information to its corresponding class token, and aggregating a group of class tokens to the consensus representation.  In addition, an image-specific dispersion module is put forward. It first refines the consensus vector with cross-attention to add more context details to the consensus. Then it performs dispersion through several Transformer blocks. Interactions within the group are also introduced in this Transformer, so the easy cases may help the difficult ones to achieve a proper dispersion result. Besides, we leverage a pre-trained Vision Transformer~\cite{ViT, CLIP} as the encoder to provide high-quality features, and an FPN-like network to decode the co-saliency masks.

In addition to the four-stage main branch for co-saliency masks generation, recent CoSOD methods tend to introduce auxiliary tasks or side pathways to enhance performance, such as additional classification task~\cite{GCoNet, DeepACG, UFO, GCoNet+}, salient object detection (SOD) priors~\cite{ICNet, CoADNet, rethink}, various kinds of contrastive learning~\cite{GCoNet, GCoNet+, DCFM, MCCL}, discriminator~\cite{MCCL} and so on. These techniques contribute to the gain in the final performance, but at the cost of complicated model structure and abundant hyper-parameters. Moreover, some of them require extra supervision in the training phase (\emph{e.g.} class label~\cite{GCoNet, DeepACG, UFO, GCoNet+}, single image saliency maps (SISM)~\cite{ICNet}). In contrast to this trend, our proposed model does not include any additional designs beyond the mask generation pipeline. Experiments show that our compact model with consensus extraction and dispersion is able to achieve comparable or superior performance compared to the previous models with complex modules.

To sum up, our contributions are as follows:

\begin{itemize}[leftmargin=*]
\item We design a hierarchical consensus extraction module that leverages high-level semantics embedded in the class token to obtain a comprehensive representation of the common object category, thus avoiding the ambiguity and incompleteness brought by using local information as consensus. It also includes iterative refinements to gradually improve the image semantics and the consensus representation.

\item This work proposes a Transformer-based dispersion module that distributes the consensus to the image feature maps in an image-specific manner, taking the diversity of the common object in different images into account. It also enables intra-group interactions during the dispersion process, thus maintaining a balance between the commonality and the specificity.

\item The two novel modules are combined with a plain ViT encoder and an FPN-like decoder to form an end-to-end trainable CoSOD network. Without additional structure and auxiliary loss, the proposed method is characterized by its simplicity and clarity. Extensive experiments on several public datasets have proven the effectiveness of our model.
\end{itemize}

\section{Related Works}
\label{sec2}

Most of the deep-learning-based CoSOD methods can be categorized into the four-stage pipeline, namely image encoding, consensus extraction, consensus dispersion, and mask decoding. A brief summary of each of above ingredients is presented below.

\paragraph{Encoder and Decoder.} For the encoder, a large body of prior works use VGG~\cite{VGG} as a defaulted choice. More recent research has explored the use of the Transformer architecture. For instance, \cite{TCNet} uses a T2T-ViT~\cite{T2T-ViT} branch accompanied with a CNN branch. \cite{MCCL} uses a single PVTv2~\cite{PVTv2}. \cite{RGBD} utilizes a Swin Transformer~\cite{Swin}. As for the decoder, the idea of feature pyramid~\cite{FPN} is often utilized. Multi-scale feature maps generated by the encoder are sent to the decoder, and the predicted masks are obtained by aggregating the maps of different scales. Although some improved designs may be introduced to these two modules (\emph{e.g.} Transformer necks between the encoder and the decoder~\cite{UFO}, decoder with intra-group interactions~\cite{CoADNet}, decoder that explicitly focuses on edges~\cite{DeepACG}), we prioritize the modules directly related to the group consensus. Our proposed network uses a plain ViT~\cite{ViT} as the encoder to get high-quality image feature map and semantic vector (\emph{i.e.} the class token), and a simple CNN-based FPN-like network as the decoder to better capture the local information such as boundaries. Such methodology decouples the task-specific designs for CoSOD with the general feature extractor and dense predictor.

\paragraph{Consensus Extraction and Dispersion.} These two components are the key to solving CoSOD, and represent the core of the network. Therefore, previous studies have proposed various implementations. As for consensus extraction, the key lies in integrating the features of a group of images. \cite{IJCAI17} concatenates the feature maps of the group and performs convolutional operations on them to get consensus. \cite{RNN} uses an RNN to sequentially read the feature map of each image, and the hidden state is used as consensus after the whole group is processed. \cite{FCN, rethink} adopt PCA to get an optimal direction (which can be seen as some type of consensus representation), and project the image feature maps to it. \cite{AAAI19} uses Low-Rank Bilinear Pooling~\cite{LRBP} to obtain the consensus vector. \cite{CoADNet} first fuses each image's SISM prediction with its feature map, then performs consensus extraction by a Group-Attentional Semantic Aggregation module that includes concatenation, block shuffle, atrous convolution, and self-attention. \cite{GICD, UFO} apply global average pooling (GAP) to the group of feature maps to get the consensus representation. \cite{GCN} and \cite{DeepACG} also employ GAP, but prior to that, they process the group of feature maps using GCN and Gromov-Wasserstein-distance-based matching~\cite{GW}, respectively. \cite{CoRP} first predicts SISMs and performs GAP to the salient regions. Then, $K$ pixels with the highest similarity to the pooled feature within the group are selected and concatenated as the consensus. \cite{GCoNet, DCFM, TCNet, GCoNet+, MCCL} utilize the attention mechanism to compute correspondences between each pair of pixels in the group. The attention matrix is then processed with pooling and softmax operation and becomes a per-pixel weight, which is used to perform weighted average pooling on the image feature maps and obtain the consensus representation.

The dispersion of the consensus representation to the image feature maps is typically defined as concatenation, addition, element-wise multiplication, dot-product, or their combination~\cite{IJCAI17, RNN, FCN, AAAI19, GICD, GCN, GCoNet, CoSFormer, DeepACG, DCFM, UFO, TCNet, GCoNet+, CoRP, MCCL}. On this basis, \cite{GICD} computes the gradient of the dot-product results with respect to the feature maps, and multiplies the pooled gradient maps to the feature maps; \cite{GCN} utilizes weighted K-means algorithm to refine the dot-product results; \cite{DCFM, TCNet} first multiplies the consensus to the feature maps, then enhances the fused maps using self-attention. \cite{CoADNet, ICNet, CADC} notice the importance of image-specific dispersion, which is similar to our ideas. Specifically, \cite{CoADNet} multiplies an element-wise weight to the consensus before adding it to the feature map of each image, and the weight is computed by a squeeze-and-excitation block. \cite{ICNet} does not produce an explicit consensus representation. Instead, it extracts salient object representations for each image separately under the guidance of SISMs. Cosine similarity is calculated between each image feature map and each representation vector, so $N$ masks are obtained for each of the $N$ images ($N$ is the size of the group). Next, it integrates the $N$ masks for each image through weighted averaging (weights are determined by the similarity between them), and the resulting single mask is multiplied with the image feature map. \cite{CADC} uses convolution and self-attention operations to transform the group of feature maps into a series of kernels (including both image-specific and group-shared ones), and implements the dispersion process by dynamic convolution.

Different from the above approaches, we put forward two novel Transformer-based modules to perform extraction at the semantic level, and implement image-specific dispersion while maintaining a consistent consensus representation.

\paragraph{Extra Modules.} Additionally, many prior works have also incorporated various additional modules or auxiliary tasks to further enhance the performance and stabilize the training. \cite{RNN} draws inspiration from the perceptual loss in neural style transfer~\cite{NST} to ensure the consistency of the predicted foreground and the ground-truth. \cite{FCN, rethink} require complicated refinements to be applied to the model output. ~\cite{GICD, CADC} design novel data augmentation approaches for training images. \cite{CoADNet, CoSFormer, CoRP} co-train the CoSOD model with an SOD branch. \cite{ICNet, rethink} use SISMs predicted by another SOD model. \cite{MCCL} adds a discriminator at the end of the pipeline to perform adversarial learning. To enhance the consensus learning, \cite{AAAI19, GCoNet, DeepACG, UFO, GCoNet+} employ the image category labels and carry out an additional classification task on the extracted consensus; \cite{GCoNet, GCoNet+} utilize contrastive learning between different groups while \cite{MCCL} further proposes to use memory-based contrastive learning; \cite{DCFM} explores self-contrastive learning by masking the co-salient object or the remaining area; \cite{CoRP} involves iterative refinement and purification.

Instead of investing efforts in this regard, we argue that these additional designs are not a necessity. The pipeline of our model is simple and direct, requiring only a minimal number of hyper-parameters and loss functions.

\begin{figure*}[t]
  \centering
  \includegraphics[width=0.8\linewidth]{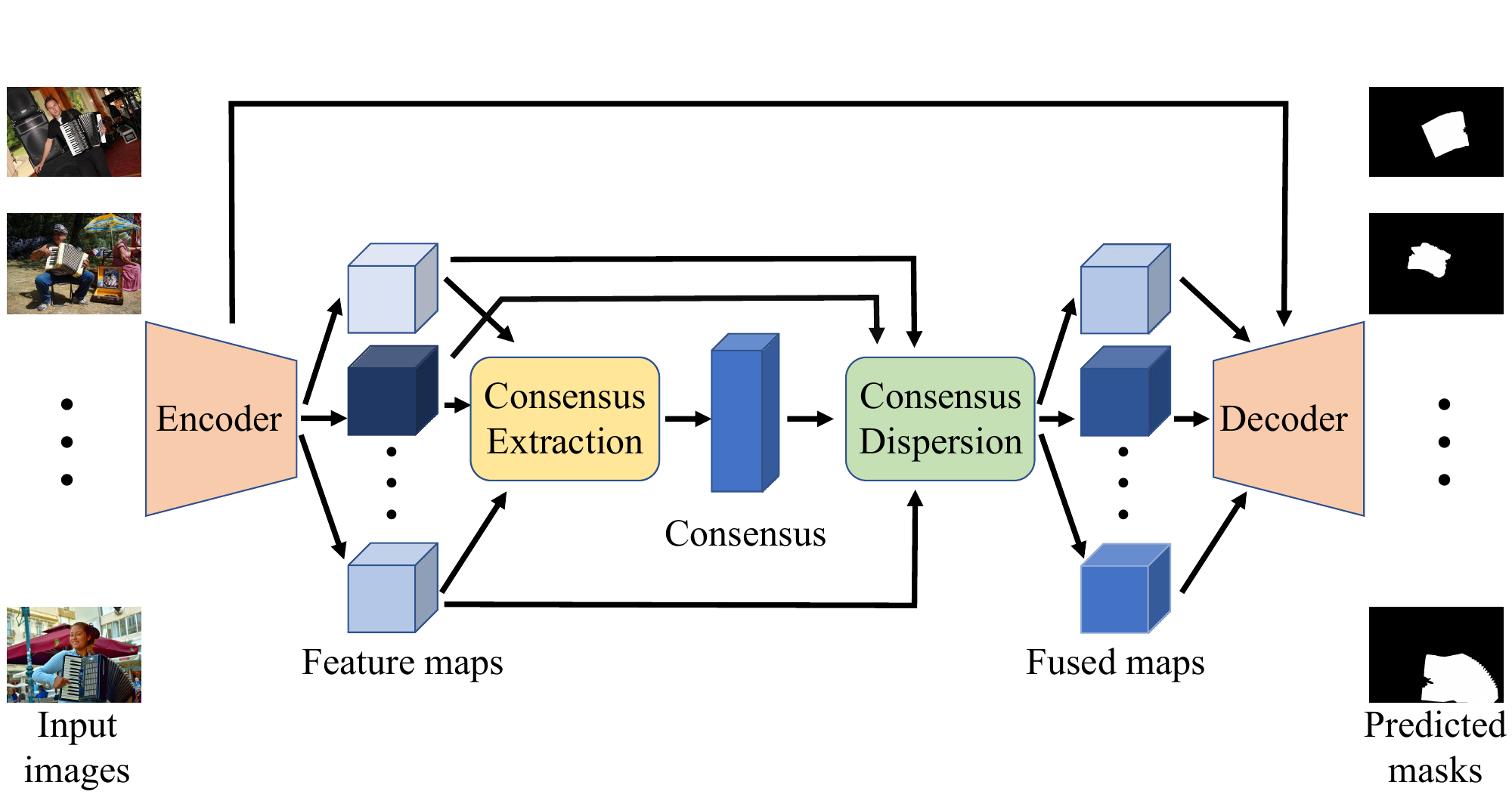}
  \caption{The pipeline of the proposed co-salient object detection model.}
  \label{fig:main}
  \Description{Pipeline}
\end{figure*}

\paragraph{Vision Transformer.} Originally proposed in the field of NLP, Transformer~\cite{Transformer} leverages the attention mechanism to process sequential data. Its notable application in computer vision, Vision Transformer (ViT)~\cite{ViT}, has shown superior performance to conventional CNN due to its ability to model global correspondence in the image. Afterward, a large number of ViT variants and related techniques have been proposed, such as \cite{Swin, PVT, PVTv2, MViT, MViTv2, T2T-ViT, BEiT, MAE}. They focus on introducing more inductive bias, improving the scalability, and designing new pre-training schemes. Though more advanced architecture may further boost the performance, we use the original ViT as backbone to keep the model simple. Such choice is also coincident with the philosophy of \cite{ViTDet, adapter} that seeks to decouple ViT pre-training and downstream fine-tuning.

Apart from the encoder, Transformer blocks are also employed in the consensus extraction and dispersion stages of our model. In this aspect, \cite{CoSFormer} shares a similar idea with our method. It simply concatenates a group of feature maps together and feeds them into a Transformer to obtain the consensus. Then it concatenates the consensus with the feature map of each image and feeds them into another Transformer for dispersion. Such an approach does not explicitly reflect the meaning of consensus, and incurs significant computation and memory overhead.  In contrast, our model leverages the high-level semantics embedded in the class token to effectively obtain a consensus representation that is better aligned  with the co-object's category. 

 In addition, several successful attempts have also been made to introduce Transformer to SOD (\emph{e.g.} \cite{VST, TriTransNet}), and RGB-D CoSOD (\emph{e.g.} \cite{RGBD}). Our proposed CoSOD method differs from these works in that it places greater emphasis on inter-image interactions without leveraging the depth information.

\section{Method}

The goal of CoSOD can be formalized as follows: given a group of images $\left\{I^n\right\}_{n=1}^N$ that contain objects of the same category, a group of co-saliency masks $\left\{M^n\right\}_{n=1}^N$ should be generated that highlight the co-salient object in each image. Here the input images are resized to the same shape, so $I^n\in\mathbb{R}^{3\times H\times W}$, while $M^n\in[0,1]^{1\times H\times W}$.

The overall pipeline of our proposed network is presented in Figure~\ref{fig:main}. As stated in Section~\ref{sec1}, four stages are involved in the network. The $L$-layer ViT encoder first takes $\left\{I^n\right\}_{n=1}^N$ as the input and generates feature maps\footnote{``Feature map'' (in $\mathbb{R}^{C\times h\times w}$) is commonly used in the context of CNN, while ``patch tokens'' (in $\mathbb{R}^{hw\times C}$) are suitable for the context of ViT. They can be converted to each other through the operations of flattening and reshaping. In this paper, we consider these two terms interchangeably.} $\left\{F^{L,n}\right\}_{n=1}^N$, $F^{L,n}\in\mathbb{R}^{C\times h\times w}$. Then the consensus extraction module extracts the representation of the common object class $\boldsymbol{g}\in\mathbb{R}^D$. The consensus dispersion module applies $\boldsymbol{g}$ to each feature map, and finally the decoder takes the fused maps and outputs mask predictions $\left\{M^n\right\}_{n=1}^N$. It should be noted that the above pipeline is also a general procedure in most of the previous works, though many of them introduce some additional branches or tasks to it.

The whole model is trained in an end-to-end fashion. Given the predictions $\left\{M^n\right\}_{n=1}^N$ and the ground-truth masks $\left\{G^n\right\}_{n=1}^N$, the objective is to minimize the Binary Cross Entropy (BCE) loss
\begin{equation}
  L_{\text{BCE}}=-\frac{1}{N}\sum_{n=1}^N\sum_{i,j}(G^n_{i,j}\log M^n_{i,j}+(1-G^n_{i,j})\log (1-M^n_{i,j})),
\end{equation}
and Intersection over Union (IoU) loss
\begin{equation}
  L_{\text{IoU}}=1-\frac{1}{N}\sum_{n=1}^N\frac{\sum_{i,j}G^n_{i,j}M^n_{i,j}}{\sum_{i,j}(G^n_{i,j}+M^n_{i,j}-G^n_{i,j}M^n_{i,j})}.
\end{equation}
Putting these two together we have
\begin{equation}
  L_{\text{total}}=L_{\text{BCE}}+L_{\text{IoU}}.
\end{equation}
The utilization of these loss functions is a common practice in CoSOD methods~\cite{DeepACG, UFO, GCoNet+, TCNet, MCCL}.

In the following subsections, we first describe the encoder and decoder in details. Based on the ViT feature, a simple yet effective baseline model is proposed. Next, we introduce the newly-designed extraction and dispersion modules, and demonstrate how to incorporate them into the baseline model to further enhance the performance.

\subsection{Encoder}
We use a plain ViT~\cite{ViT} as the image encoder. The input image is first divided into non-overlapping patches and linearly projected to obtain a sequence of token embeddings. These patch tokens along with a learnable class token are then fed into multiple Transformer layers. The Transformer architecture enables efficient parallel processing and effective global interactions.

Previously, CoSOD models that use Transformer-base encoders typically employ variants of ViT that are better suited for dense prediction tasks, \emph{e.g.} PVTv2~\cite{PVTv2} in~\cite{MCCL}, T2T-ViT~\cite{T2T-ViT} in \cite{TCNet}, Swin~\cite{Swin} in \cite{RGBD}. On the contrary, the use of original ViT decouples the design of the pre-training process and the need of downstream tasks, as mentioned in~\cite{ViTDet, adapter}. Therefore, our model is fully compatible with the research progress in pre-training, scalability, and other aspects of ViT.
In this paper, we choose the ViT model pre-trained with CLIP~\cite{CLIP}. CLIP adds a linear projection to the original ViT's output, and the projected class token is treated as the CLIP image feature vector. Meanwhile, CLIP utilizes another Transformer to encode the sentence corresponding to the image into a CLIP text vector, and performs cross-modal contrastive learning.
We choose CLIP's ViT weight for two reasons.
Firstly, the large-scale contrastive pre-training of CLIP will endow the image features with rich and high-quality semantic information. Secondly, the sentence supervision used by CLIP provides a more complete description of the images compared to other supervision signals such as category labels, which ensures that image features will not miss crucial information about the co-object when the scene is complex.
More advanced pre-training schemes can be easily introduced to our model by simply changing the encoder's initial weight.

\subsection{Decoder}
Following the previous works~\cite{DeepACG, UFO, GCoNet, GCoNet+, CoRP, MCCL}, we adopt an FPN-like~\cite{FPN} decoder, whose input includes not only the image feature maps fused with consensus, but also the output of intermediate layers in the encoder. Unlike CNNs and some ViT variants that have hierarchical architecture, the intermediate features of ViT remain at the same scale throughout the network, and the output of earlier layers does not necessarily contain finer-grained information than that of later layers. Thus, we ignore the order of these intermediate maps and simply sum them together. We do not use the multi-scale supervision like~\cite{GICD, ICNet, CADC, GCoNet, DCFM, GCoNet+, MCCL} either, \emph{i.e.} the loss is computed only once. Specifically, the decoder works as:

\begin{equation}
 \text{Dec}(F^1,...,F^K)=\text{Conv}(\sum_{l=1}^K\text{Up}^{l}(F^l)),
\end{equation}
where $\text{Conv}$ is a $1\times1$ convolution that reduces the channel number to one and produces the mask prediction; $\text{Up}^{l}$ is a series of $3\times3$ convolutions and $2\times$ bilinear upsamplings that progressively expand the spatial dimension.

\begin{figure*}[h]
  \centering
  \includegraphics[width=\linewidth]{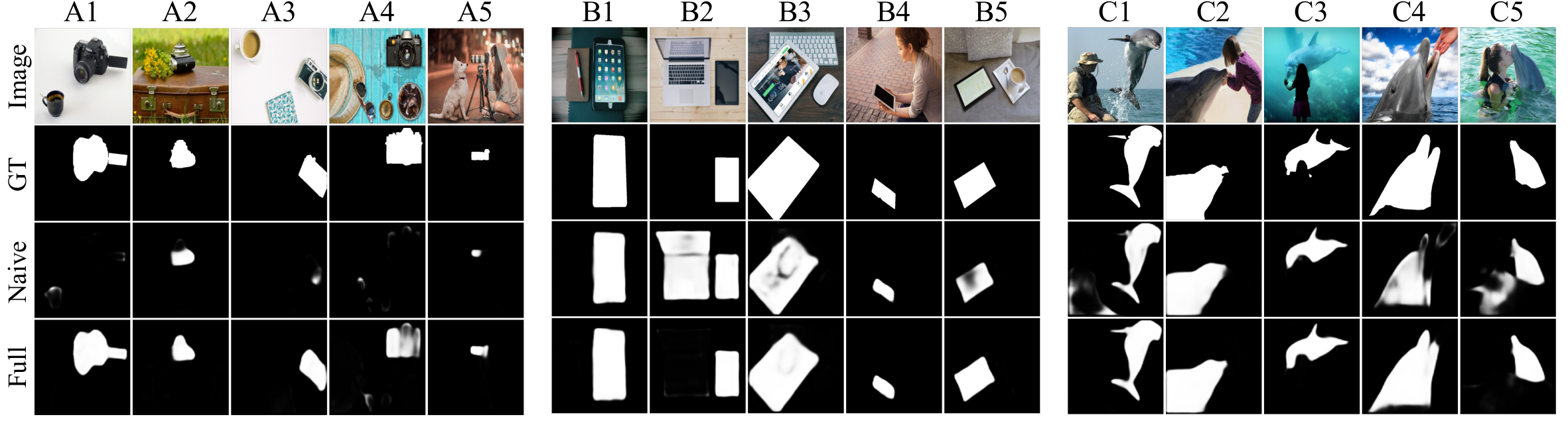}
  \caption{Failure cases of the naive approach. A/B/C represent the group of camera/tablet/dolphin, respectively. The input images, the ground-truth masks, the results of our naive model and full model are presented. In group A, the naive model cannot capture the target object in many cases. While in group B/C, the majority of the results are correct (\emph{e.g.} B1/B4/C2/C3), but some predictions fail to fully cover the details of the co-object (B3/B5/C1/C4) or are interfered by other objects (B2/C1/C5).}
  \label{fig:try}
  \Description{Failure Cases}
\end{figure*}

\subsection{A Naive Approach}
\label{subsec:naive}
Based on the pre-trained encoder, a straightforward idea is to extract consensus from the CLIP feature vectors. As an initial try, we simply use the mean of these vectors after normalization as consensus, and disperse it to the image feature maps by element-wise multiplication. Formally, we first 
 obtain the CLIP feature vectors:
\begin{equation}
 \boldsymbol{v}_{\text{CLIP}}^n=\text{Linear}_{\text{CLIP}}(c^{L,n}),n=1,...,N,
\end{equation}
where $c^{L,n}$ is the class token in the ViT's final output, $\text{Linear}_{\text{CLIP}}$ is CLIP's additional linear projection. Next, we conduct the averaging operation:
\begin{equation}
\label{equ:extraction-avg}
 \boldsymbol{g}=\frac{1}{N}\sum_{n=1}^N\text{Norm}(\boldsymbol{\boldsymbol{v}_{\text{CLIP}}^n}),
\end{equation}
where $\text{Norm}$ is the $l2$ normalization. And
\begin{equation}
\label{equ:dispersion-mul}
 \hat{F}^{l,n}=F^{l,n}\cdot\text{Linear}_{\text{Dis}}^{l}(\boldsymbol{g}), l\in\left\{l_1,...,l_K\right\}, n=1,...,N,
\end{equation}
\begin{equation}
 M^n=\text{Dec}(\hat{F}^{l_1,n},...,\hat{F}^{l_K,n}), n=1,...,N,
\end{equation}
where $F^{l,n}$ is the output feature map of ViT's layer $l$; $\text{Linear}_{\text{Dis}}$ is a linear projection that adjusts the dimension of the consensus vector $\boldsymbol{g}$; $\hat{F}^{l,n}$ is the map after dispersion; $\left\{l_1,...,l_K\right\}\subset\left\{1,...,L\right\}$ are the selected intermediate layers' indices ($l_K=L$). 

Note that in (\ref{equ:dispersion-mul}), we disperse $\boldsymbol{g}$ to not only $F^{L,n}$ but also all the inputs of the decoder, so as to reinforce the consensus signal during the decoding process. This multi-stage dispersion is also a commonly adopted procedure in prior works~\cite{AAAI19, DeepACG, UFO}.

It is shocking that such a simple strategy has achieved a fairly competitive performance compared with previous works, as can be later seen in Table~\ref{tab:res}. The results demonstrate the quality of CLIP vectors and the effectiveness of the four-stage pipeline. In order to further improve the model, we examine the output masks, and some typical results are shown in Figure~\ref{fig:try}. Two types of problems can be identified. (I) In the first case, the predictions of \emph{many} images in the group deviate significantly from the target object, indicating that there may be further room for improvement in the consensus extraction stage. (II) In the second case, most images' co-object is detected correctly, while \emph{a few} hard examples make the model go wrong. It can be inferred that the consensus is well-extracted in these groups, but it is not sufficient to guide the model to capture the whole co-object or distinguish the co-object from interference in a complex scene.

\subsection{Semantic-Level Consensus Extraction}
We first modify Equation (\ref{equ:extraction-avg}) to facilitate more advanced consensus extraction. In addition to the CLIP vectors, we also utilize the CLIP feature maps to include richer information, \emph{i.e.}
\begin{equation}
 F^{L,n}_{\text{CLIP}}=\text{Linear}_{\text{CLIP}}(F^{L,n}),n=1,...,N.
\end{equation}

\begin{figure}[t]
  \centering
  \includegraphics[width=\linewidth]{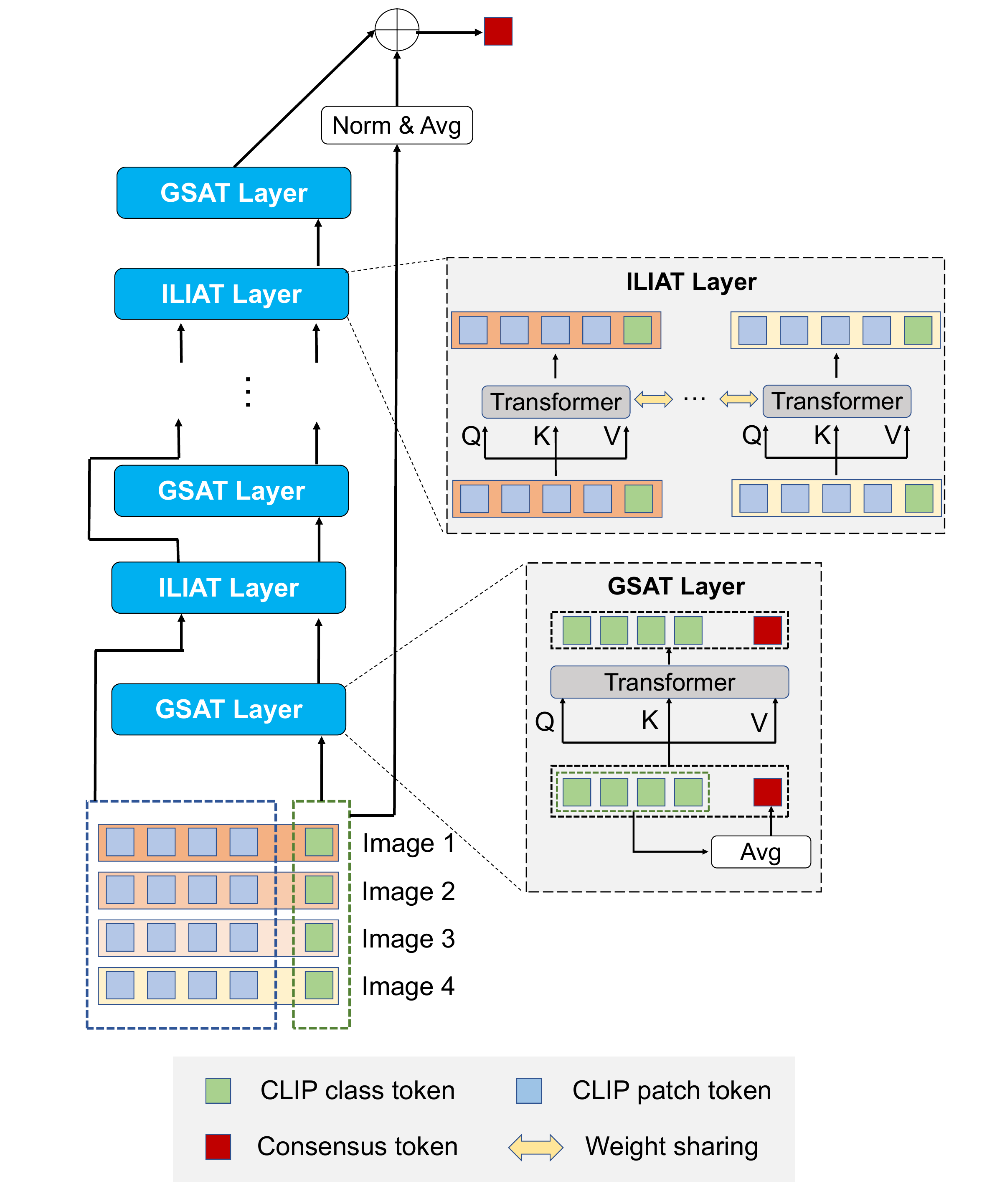}
  \caption{The proposed consensus extraction module. As for the ILIAT layer, a self-attention-based version is presented here, which can be replaced by a cross-attention-based version as in Equation (\ref{equ:extraction-ILIAT-CA}).}
  \label{fig:extraction}
  \Description{Extraction}
\end{figure}

Based on the feature maps $\left\{F^{L,n}_{\text{CLIP}}\right\}_{n=1}^N$ and the semantic vectors $\left\{\boldsymbol{v}_{\text{CLIP}}^n\right\}_{n=1}^N$, two steps are performed alternatively, as shown in Figure~\ref{fig:extraction}. In the global step, a group-based semantic agreement Transformer (GSAT) layer refines the image semantics and extracts consensus simultaneously. Specifically, it takes the semantic vectors along with their average as input and performs self-attention:
\begin{equation}
 [\bar{\boldsymbol{v}}^{(i)},\left\{\boldsymbol{v}^{n,(i)}_g\right\}_{n=1}^N]=\text{Trans}_{\text{GSAT}}^{(i)}([\frac{1}{N}\sum_{n=1}^N\boldsymbol{v}^{n,(i)},\left\{\boldsymbol{v}^{n,(i)}\right\}_{n=1}^N]),
\end{equation}
where $i=1,...,M+1$ is the layer index; the initial $\boldsymbol{v}^{n,(1)}$ is $\boldsymbol{v}_{\text{CLIP}}^n$. In the local step, an image-based local information aggregation Transformer (ILIAT) layer comes into play. It processes each image individually, and can be defined with self-attention: 
\begin{equation}
 [\boldsymbol{v}^{n,(i+1)},F^{L,n,(i+1)}_{\text{CLIP}}]=\text{Trans}_{\text{ILIAT-SA}}^{(i)}([\boldsymbol{v}^{n,(i)}_g,F^{L,n,(i)}_{\text{CLIP}}]),
\end{equation}
or cross-attention:
\begin{equation}
\label{equ:extraction-ILIAT-CA}
 \boldsymbol{v}^{n,(i+1)}=\text{Trans}_{\text{ILIAT-CA}}^{(i)}(\boldsymbol{v}^{n,(i)}_g,F^{L,n}_{\text{CLIP}},F^{L,n}_{\text{CLIP}}),
\end{equation}
where $i=1,...,M; n=1,...,N$; the three inputs of $\text{Trans}_{\text{ILIAT-CA}}$ are used as query, key, and value, respectively; the initial $F^{L,n,(1)}_{\text{CLIP}}$is $F^{L,n}_{\text{CLIP}}$. 

Intuitively, ILIAT processes local information and aggregates the salient object features of each individual image into its semantic vector. The GSAT layer facilitates semantic-level information exchange, extracting the common component from the semantics of individual images. It is similar in effect to the averaging operation in Equation (\ref{equ:extraction-avg}), but the additional parameters and self-attention mechanism allow for more flexible interactions of the image semantics. GSAT also updates the semantic vector of each image, thus they can participate in the following ILIAT layer and provide more reliable guidance for the aggregation of local information.

After $M+1$ layers of GSAT and $M$ layers of ILIAT, the last GSAT's output corresponding to the average semantic token is treated as the final consensus representation. To fully utilize the important semantic priors endowed by the pre-training of CLIP vectors, we also add Equation (\ref{equ:extraction-avg}) to the consensus, \emph{i.e.},
\begin{equation}
 \boldsymbol{g}=\bar{\boldsymbol{v}}^{(M+1)}+\frac{1}{N}\sum_{n=1}^N\text{Norm}(\boldsymbol{\boldsymbol{v}_{\text{CLIP}}^n}).
\end{equation}

The proposed Transformer-based consensus extraction differs from the commonly used attention-based method~\cite{GCoNet, DCFM, TCNet, GCoNet+, MCCL} in that it leverages the ViT's class token to perform semantic-level aggregation and iterative refinement. Therefore, it is able to grasp the complete representation of the co-object without being easily disrupted by other objects that are locally similar to the target. It also reduces the cost of computing the attention matrix. More detailed comparisons can be found in the supplementary materials.

\subsection{Image-Specific Consensus Dispersion}
\begin{figure}[t]
  \centering
  \includegraphics[width=\linewidth]{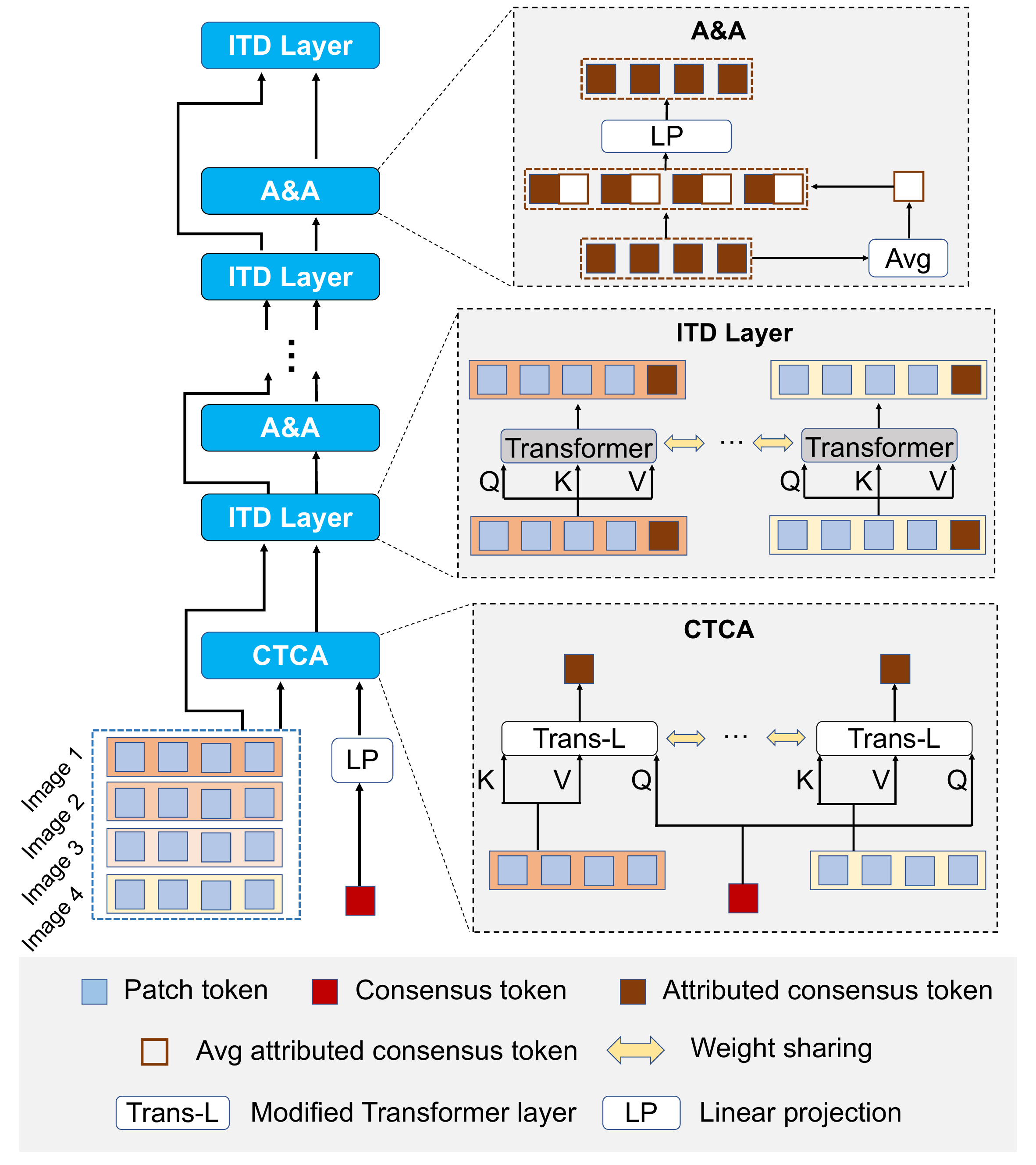}
  \caption{The proposed dispersion module.}
  \label{fig:dispersion}
  \Description{Dispersion}
\end{figure}

In order to better cope with the variation of the co-object, we design another Transformer-based module to perform the dispersion of consensus to image features, as shown in Figure~\ref{fig:dispersion}. It first situates the consensus representation within the specific context of each image: 
\begin{equation}
\label{equ:dispersion-CTCA}
 \tilde{\boldsymbol{g}}^n=\text{Trans}_{\text{CTCA}}(\text{Linear}_{\text{CTCA}}(\boldsymbol{g}), F^{L,n}, F^{L,n}),n=1,...,N,
\end{equation}
where $\text{Trans}_{\text{CTCA}}$ is the contextualization Transformer layer with cross-attention (CTCA), and the three inputs are used as query, key, and value, respectively; $\text{Linear}_{\text{CTCA}}$ is a linear projection that adjusts the channel number. The output $\tilde{\boldsymbol{g}}^n$ thus represents the image-specific status and attributions of the co-object in image $I_n$. There is a slight difference between $\text{Trans}_{\text{CTCA}}$ and the original Transformer layer. We multiply the results of cross-attention and MLP with a learnable channel-wise coefficient, thus allowing the network to adaptively learn the degree of contextualization.

Subsequently, the attributed consensus is concatenated with the corresponding image feature map and processed by multiple image-based Transformer layers for dispersion (ITD). Moreover, to further facilitate interactions within the group, we insert an aggregation and allocation (A\&A) operation between adjacent ITD layers. It is implemented by concatenating the output consensus token with their average and applying a linear projection, as shown below in Equation (\ref{equ:dispersion-AA}). Through the communication of the attributed consensus, the easy cases (where the co-object can be effortlessly located) may help the difficult cases (where the co-object is concealed or interference exists) to better focus on the common object. Besides, the A\&A process guarantees that the dispersion will not deviate significantly from the extracted consensus while ensuring the flexibility of the dispersion (\emph{i.e.} multiple Transformer layers). A similar idea is also adopted by~\cite{CoADNet} in its decoder design.

Besides, we also retain the dispersion implemented by multiplication, since it has exhibited satisfactory performance and remains fully compatible with the improved pipeline. To sum up, Equation (\ref{equ:dispersion-mul}) in the naive approach is modified to Equation (\ref{equ:dispersion-CTCA}) followed by:
\begin{equation}
\left \{
    \begin{array}{l}
        [\tilde{\boldsymbol{g}}^{n,(i+1)}_{\text{pre}},F^{L,n,(i+1)}]=\text{Trans}_{\text{ITD}}^{(i)}([\tilde{\boldsymbol{g}}^{n,(i)},F^{L,n,(i)}])\\
        \tilde{\boldsymbol{g}}^{n,(i+1)}=\text{Linear}_{\text{A\&A}}^{(i)}([\tilde{\boldsymbol{g}}_{\text{pre}}^{n,(i+1)},\frac{1}{N}\sum_{n=1}^N\tilde{\boldsymbol{g}}_{\text{pre}}^{n,(i+1)}]) \label{equ:dispersion-AA}\\
    \end{array}
\right. ,
\end{equation}
\begin{equation}
 \hat{F}^{l,n}=F^{l,n}\cdot\text{Linear}_{\text{Dis}}^{l}(\tilde{\boldsymbol{g}}_{\text{pre}}^{n,(M'+1)}),
\end{equation}
\begin{equation}
 \hat{F}^{L,n}=F^{L,n,(M'+1)}\cdot\text{Linear}_{\text{Dis}}^{L}(\tilde{\boldsymbol{g}}_{\text{pre}}^{n,(M'+1)}),
\end{equation}
where $l\in\left\{l_1,...,l_{K-1}\right\}; n=1,...,N; i=1,...,M'$ is the layer index; the initial $ \tilde{\boldsymbol{g}}^{n,(1)}=\tilde{\boldsymbol{g}}^{n}, F^{L,n,(1)}=F^{L,n}$.

\begin{table*}[h]
  \caption{Performance comparision with the state-of-the-art methods. C/S/D stand for COCO-9213/COCO-SEG/DUTS-class dataset, respectively. 
  The best and second-best results under each metric are marked in bold and underlined respectively.}
  \label{tab:res}
  \begin{center}
\renewcommand{\arraystretch}{1.0}
\renewcommand{\tabcolsep}{1.14mm}
\begin{tabular}{r|c|cccc|cccc|cccc}
\hline
& Training & \multicolumn{4}{c|}{CoCA~\cite{GICD}} & \multicolumn{4}{c|}{CoSOD3k~\cite{rethink}} & \multicolumn{4}{c}{CoSal2015~\cite{IJCV16}} \\
Method & Set & $E_{\xi}^\mathrm{max} \uparrow$ & $S_\alpha \uparrow$ & $F_\beta^\mathrm{max} \uparrow$ & $\epsilon \downarrow$ & $E_{\xi}^\mathrm{max} \uparrow$ & $S_\alpha \uparrow$ & $F_\beta^\mathrm{max} \uparrow$ & $\epsilon \downarrow$ & $E_{\xi}^\mathrm{max} \uparrow$ & $S_\alpha \uparrow$ & $F_\beta^\mathrm{max} \uparrow$ & $\epsilon \downarrow$ \\
\hline
DeepACG(CVPR21)\cite{DeepACG} & S & 0.771 & 0.688 & 0.552 & 0.102 & 0.838 & 0.792 & 0.756 & 0.089 & 0.892 & 0.854 & 0.842 & 0.064 \\
GCoNet(CVPR21)\cite{GCoNet} & D & 0.760 & 0.673 & 0.544 & 0.105 & 0.860 & 0.802 & 0.777 & 0.071 & 0.887 & 0.845 & 0.847 & 0.068 \\
CoEGNet(TPAMI21)\cite{rethink} & D & 0.717 & 0.612 & 0.493 & 0.106 & 0.825 & 0.762 & 0.736 & 0.092 & 0.882 & 0.836 & 0.832 & 0.077 \\
CADC(ICCV21)\cite{CADC} & C+D & 0.744 & 0.681 & 0.548 & 0.132 & 0.840 & 0.801 & 0.759 & 0.096 & 0.906 & 0.866 & 0.862 & 0.064 \\
CoSFormer(arxiv21)\cite{CoSFormer} & C+D & 0.770 & 0.724 & 0.603 & 0.103 & 0.879 & 0.835 & 0.807 & 0.066 & 0.929 & {\bfseries 0.894} & 0.891 & {\bfseries 0.047}\\
DCFM(CVPR22)\cite{DCFM} & C & 0.783 & 0.710 & 0.598 & 0.085 & 0.874 & 0.810 & 0.805 & 0.067 & 0.892 & 0.838 & 0.856 & 0.067 \\
UFO(TMM23)\cite{UFO} & S & 0.782 & 0.697 & 0.571 & 0.095 & 0.874 & 0.819 & 0.797 & 0.073 & 0.906 & 0.860 & 0.865 & 0.064 \\
CoRP(TPAMI23)\cite{CoRP} & C+D & 0.741 & 0.703 & 0.575 & 0.110 & 0.866 & 0.825 & 0.801 & 0.072 & 0.915 & 0.877 & 0.888 & \underline{0.049} \\
GCoNet+(TPAMI23)\cite{GCoNet+} & S+D & \underline{0.814} & \underline{0.738} & {\bfseries 0.637} & \underline{0.081} & 0.901 & 0.843 & 0.834 & 0.062 & 0.924 & 0.881 & 0.891 & 0.056 \\
MCCL(AAAI23)\cite{MCCL} & S+D & 0.796 & 0.714 & 0.590 & 0.103 & 0.903 & \underline{0.858} & 0.837 & 0.061 & 0.927 & \underline{0.890} & 0.891 & 0.051 \\
\hline
Ours (naive) & S+D & 0.804 & 0.707 & 0.598 & {\bfseries 0.080} & \underline{0.914} & 0.851 & \underline{0.847} & \underline{0.056} & {\bfseries 0.942} & 0.888 & {\bfseries 0.903} & {\bfseries 0.047} \\
Ours (full model) & S+D & {\bfseries 0.831} & {\bfseries 0.741} & \underline{0.631} & \underline{0.081} & {\bfseries 0.922} & {\bfseries 0.863} & {\bfseries 0.857} & {\bfseries 0.054} & \underline{0.941} & \underline{0.890} & \underline{0.902} & {\bfseries 0.047}\\
\end{tabular}
\end{center}
\end{table*}

\section{Experiments}
\label{sec4}
\subsection{Datasets}

We evaluate the proposed method on three widely-used CoSOD datasets, including Cosal2015~\cite{IJCV16}, CoSOD3k~\cite{rethink}, and CoCA~\cite{GICD}. Both of CoSOD3k and CoCA are  challenging since there are many distracting objects in each group, and the co-object exhibits high diversity in different images. The metrics for evaluation include maximum E-measure~\cite{E}, S-measure~\cite{S}, maximum F-measure~\cite{F}, and mean absolute error (MAE)~\cite{MAE_err}. We use the evaluation tools provided by GICD~\cite{GICD}. The evaluation is performed at the original scale of each image.

There are currently three supervised training sets for CoSOD, namely COCO-9213~\cite{IJCAI17}, DUTS-class~\cite{GICD}, and COCO-SEG~\cite{AAAI19}. Our model is trained on the combination of COCO-SEG and DUTS-class to keep pace with the state-of-the-art (SOTA) methods~\cite{GCoNet+, MCCL}.

\subsection{Implementation Details}
 All images are rescaled to $224\times224$ as the input. The number of GSAT/ILIAT/ITD layers are set to 3/2/4, respectively. The batch size is set to 16 for training. During inference, the whole group is processed in a single pass regardless of its size. The model is trained using Adam optimizer for 15 epochs. The learning rate is set to 3e-5.
We use the CLIP~\cite{CLIP} pre-trained ViT-B/16 as the image encoder. To take full advantage of the pre-trained weight, we set the learning rate of the encoder to $\frac{1}{10}$ of the rest of the network. 

All experiments are conducted on a single NVIDIA GeForce 3080Ti. 
Using the benchmark designed by \cite{MCCL}, the inference speed is $\sim$95fps, which is on par with recent SOTA methods.
Please refer to the appendix for a detailed description on the model configuration.

\subsection{Main Results}
\begin{figure*}[h]
  \centering
  \includegraphics[width=0.95\linewidth]{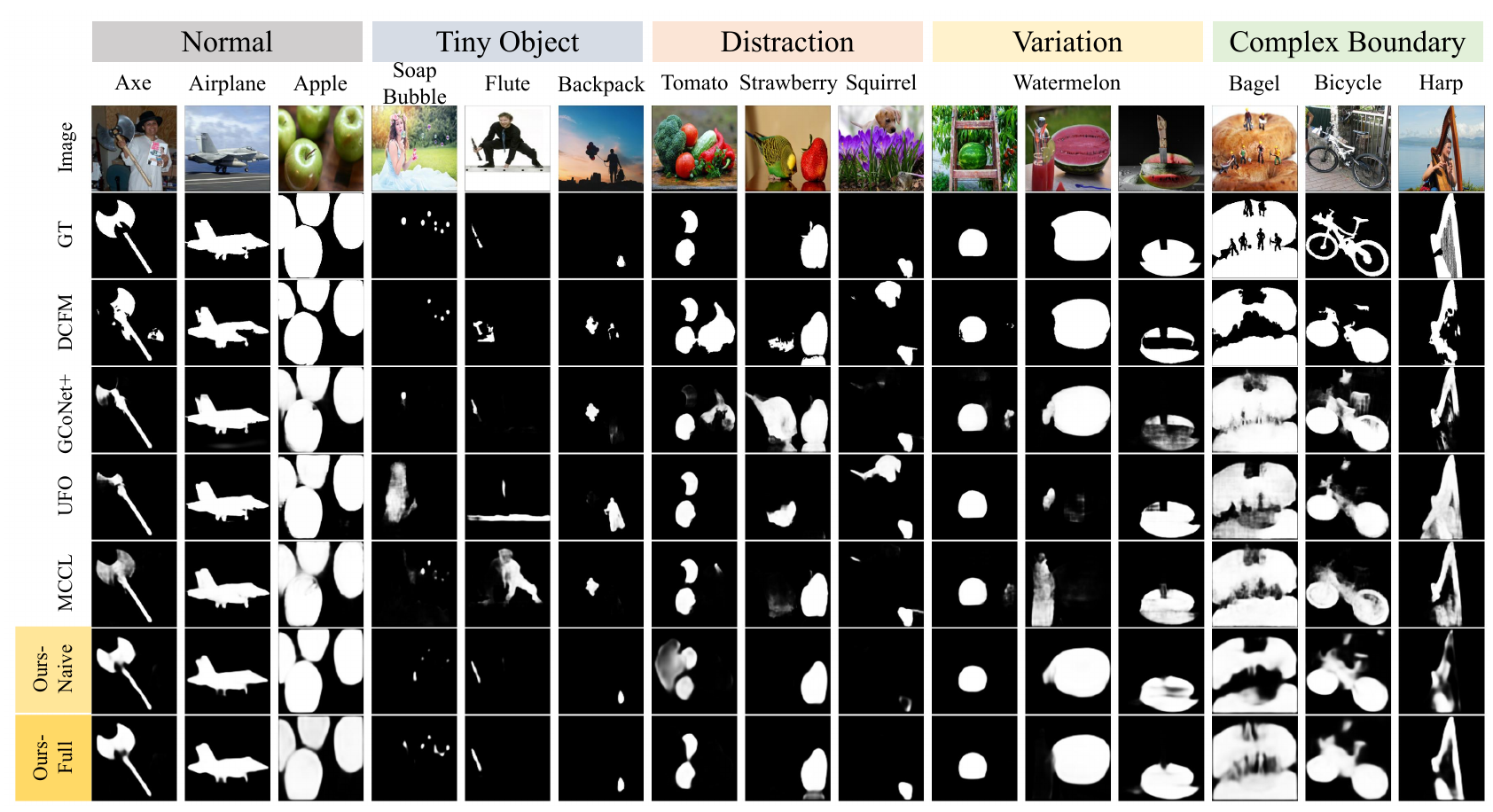}
  \caption{Qualitative results of the proposed method and some state-of-the-art methods.}
  \label{fig:quality}
  \Description{Qualitative results}
\end{figure*}

\paragraph{Quantitative Results.} Table ~\ref{tab:res} shows the quantitative performances of our model and some representative works in recent years. To reduce the effect of randomness, we independently train our model three times and report the averaged performance. Among the previous works, CoSFormer~\cite{CoSFormer}, GCoNet+~\cite{GCoNet+}, MCCL~\cite{MCCL} achieve SOTA performance on CoSal2015, CoCA, CoSOD3k, respectively. Our naive model performs better than CoSFormer on CoSal2015 and better than MCCL on CoSOD3k (except for S-measure). Its results on CoCA are similar to that of CoSFormer and MCCL but inferior to that of GCoNet+. Equipped with the newly-designed extraction and dispersion modules, our full model demonstrates better performance. It surpasses the naive model on the two hard test sets (CoSOD3k and CoCA) while maintaining competitive performance on the simple one (CoSal2015). Specifically, the performance of the full model on CoCA is comparable to that of GCoNet+, while the pipeline and objective function of our model are much simpler than GCoNet+ and MCCL. The full model performs exceptionally well on CoSOD3k, surpassing all previous methods by a large margin (\emph{e.g.} +1.9\% max E-measure and +2.0\% max F-measure). It also outperforms CoSFormer on most metrics, showing the benefits of designing task-specific Transformer modules.

\paragraph{Qualitative Results.} Figure ~\ref{fig:quality} shows some predicted masks of our model and previous SOTA methods. Five types of test images are considered here, including the normal cases and four typical kinds of difficult cases. ``Tiny Object'' requires the model not to miss the subtle details in the image. By leveraging the rich semantics in the CLIP vector and aggregating the local information in the ILIAT layer, our model manages to detect very small co-objects like bubbles. ``Distraction'' implies the existence of salient interfering objects in the image, while ``Variation'' means that the co-object exhibits very different appearances in different images. Methods that rely solely on local information may be misled by such complex scenes, resulting in detecting incorrect objects or the omission of certain parts of the co-object. In contrast, our method avoids these issues through semantic-level consensus extraction and image-specific dispersion, thus it effectively obtains the information of the co-object and conducts accurate detection. Lastly, ``Complex Boundary'' refers to the images with complicated co-object structures. Due to the lack of specialized modules for boundaries and the relatively low resolution of the network, our model does not perform perfectly on the fine structures, which is also a common problem in existing CoSOD methods. In future studies, we plan to refer to the latest research in the field of image segmentation to obtain insights for addressing this issue.

\begin{table}[t]
\small
  \caption{Impacts of the proposed modules.}
  \label{tab:modules}
  \begin{tabular}{c|c|cccc}
    \toprule
    Extraction & Dispersion & $E_{\xi}^\mathrm{max} \uparrow$ & $S_\alpha \uparrow$ & $F_\beta^\mathrm{max} \uparrow$\\
    \midrule
    avg CLIP vec& multiply & 0.804 & 0.707 & 0.598\\
    \textbf{GSAT} & multiply & 0.808 & 0.715 & 0.606\\
    \textbf{GSAT+ILIAT(CA)} & multiply & 0.806 & 0.721 & 0.610\\
    \textbf{GSAT+ILIAT(SA)} & multiply & 0.809 & 0.710 & 0.604\\    
    avg CLIP vec & \textbf{CTCA} & 0.812 & 0.715 & 0.598\\
    avg CLIP vec & \textbf{CTCA+ITD} & 0.820 & 0.733 & 0.624 \\
         & (w/o A\&A) &  \\
    avg CLIP vec & \textbf{CTCA+ITD} & 0.827 & 0.738 & 0.630\\
    \textbf{GSAT+ILIAT(CA)} & \textbf{CTCA+ITD} & {\bfseries0.831} & {\bfseries0.741} & {\bfseries0.631}\\
  \bottomrule
\end{tabular}
\end{table}

\subsection{Ablation Studies}
We conduct experiments to examine the role of the proposed modules. Different choices of the extraction and dispersion method are tried as in Table~\ref{tab:modules}. We report the results on CoCA since it contains the most challenging scenarios which are precisely the focus of our model design.
It can be concluded that: (1) each module (GSAT, ILIAT, CTCA, ITD w/ A\&A) can improve the performance of the naive method. (2) Combining these modules further enhances the performance, indicating the complementary nature of the extraction and dispersion process. (3) Using cross-attention in ILIAT exhibits better performance than using self-attention, suggesting that updating local features during consensus extraction is not necessary. (4) The improvement in performance brought by modifying the dispersion stage (especially introducing ITD) is greater than that of modifying the extraction stage, which means that designing a distribution strategy capable of handling complex scenes is more significant than improving the quality of the naive average-CLIP-vector-based consensus representation. In Section~\ref{subsec:naive}, we note that the naive model has fewer Type I failures (mainly due to extraction issues) than Type II failures (mainly due to dispersion issues), which is consistent with the experimental findings. Additional ablation experiments are provided in the supplementary materials.

\section{Conclusion}
We summarize the mainstream CoSOD methods into a general four-stage paradigm. Based on that, we note two major issues regarding the group consensus. We propose to focus on semantic-level information to extract a comprehensive consensus representation, and consider the image-specific variation of the co-object when dispersing the consensus to image features.
We first present a simple yet effective model leveraging high-level semantics in CLIP vectors. Then an improved model is put forward that involves hierarchical consensus extraction, iterative semantics refinement, image-specific consensus attribution, and dispersion with cross-image interactions. The performance of the full model and the effectiveness of its modules are demonstrated by extensive experiments.

\begin{acks}
This work is supported by Beijing Natural Science Foundation (Z190001), National Key R\&D Program of China (2022ZD0160305) and a research grant from BOE Technology.
\end{acks}

\clearpage


\bibliographystyle{ACM-Reference-Format}
\balance
\bibliography{sample-base}

\clearpage

\appendix

\section{Model Configurations}
In this section, we provide a detailed description of the specific structure and hyper-parameters choice of the proposed model.
\subsection{Encoder}
Following the ViT-B/16 in \cite{ViT}, the ViT encoder has 12 Transformer layers. The patch size is $16\times16$. The hidden dimension is 768, and the number of Transformer heads is 12. We use the pre-trained weights provided by CLIP~\cite{CLIP}. To be consistent with CLIP, the output of the ViT's final layer is processed with layer normalization and linear projection, resulting in $512d$ tokens. 

\subsection{Consensus Extraction and Dispersion}
The hidden dimension of GSAT and ILIAT is 512, which is the same as the CLIP feature dimension. In CTCA, the $512d$ consensus is first linearly projected to $768d$. Then it participates in the cross-attention with the $768d$ ViT feature maps. The hidden dimension of ITD is set to 768, which is the same as the dimension of the feature maps. 
We intuitively set the number of GSAT layers to 3, the number of ILIAT layers to 2, and the number of ITD layers to 4. The number of Transformer heads is set to 8 in all these modules. The A\&A module projects the $768d+768d$ concatenated vector to $768d$. The projection before multiplying the consensus to the feature maps is $768d\rightarrow768d$ for the full model, and $512d\rightarrow768d$ for the naive model.

\subsection{Decoder}
The indices of the intermediate features that are sent to the decoder (denoted as $\left\{l_1,..,l_K\right\}$ in the paper) are $\left\{4, 6, 8, 12\right\}$. They are all of the shape $N\times196\times768$, where $N$ is the size of the group, $196=\frac{224}{16}\times\frac{224}{16}$ is the number of tokens (the class token is discarded). In the full model, ITD is applied to the last group of features (\emph{i.e.} index 12) and it does not change the shape. In the decoder, each group of features is first reshaped to $N\times768\times14\times14$ and then processed by three consecutive CNN blocks. Each block contains a $3\times3$ convolution followed by batch normalization, ReLU activation, and a $2\times$ bilinear upsampling in sequence. The first block reduces the channel number to 256, while the following blocks do not change the channel dimension. After that, the four groups of features with shape $N\times256\times112\times112$ are added together. The sum is sent to another $3\times3$ convolution and bilinearly upsampled again to get the final output.

\subsection{Training Scheme}
The code is implemented with PyTorch. During training, the batch size is set to 16 and we use the batch sampler designed by \cite{ICNet}. In each iteration, it randomly selects a group and samples 16 images from it. Batches with less than 16 images are allowed, but batches with only one image are discarded since CoSOD is ill-defined for a single image. No image will be sampled more than once in one epoch. During inference, each group is processed in a single pass, regardless of the group size. The input image is resized to $224\times224$ and normalized for both training and inference. The ground-truth is resized to $224\times224$ for training, and the loss is calculated at this scale. When evaluating the performance at the inference stage, the network output is resized to the image's original size. We use the evaluation tools provided by ~\cite{GICD}\footnote{https://github.com/zzhanghub/eval-co-sod}. No data augmentation is used during training, except for horizontal flipping with a probability of 50\%.

We combine the DUTS\_class~\cite{GICD} dataset and the COCO-SEG~\cite{AAAI19} dataset together, so there are 369 groups and 209k images in total. With the sub-group size (\emph{i.e.} batch size) set to 16, each training epoch contains 13k iterations. We train the model for 15 epochs (about 200k iterations). We use Adam optimizer with weight decay set to 1e-4 and betas set to [0.9, 0.99] empirically. The learning rate is set to 3e-5 by a few trials.

\subsection{Model Efficiency}

\begin{table}[h]
  \caption{Model statistics.}
  \label{tab:stat}
  \begin{tabular}{c|c|c|c|c}
    \toprule
    Model & \#params & FLOPs & inference time & fps\\
    \midrule
    DCFM~\cite{DCFM} & 142.3M & 31.7G & 0.008s & $\sim$125 \\
    GCoNet+~\cite{GCoNet+} & 18.4M & 27.5G & 0.009s & $\sim$120\\
    MCCL~\cite{MCCL} & 27.0M & 5.9G & 0.017s & $\sim$60\\
    DMT~\cite{DMT} & 40.4M & 84.4G & 0.023s & $\sim$45 \\
    \midrule
    Ours-naive & 99.6M & 28.4G & 0.007s & $\sim$150\\
    Ours-full & 156.7M & 34.7G & 0.010s & $\sim$95\\
  \bottomrule
\end{tabular}
\end{table}

Since CoSOD is a group-based task, the inference time is related to the size of the group. On a single NVIDIA GeForce 3080Ti, the speed of our full model is about 0.10s per group in CoCA~\cite{GICD} (about 16 images per group), 0.12s per group in CoSOD3k~\cite{rethink} (about 21 images per group), and 0.23s per group in CoSal2015~\cite{IJCV16} (about 40 images per group). The speed of the naive model is about 0.09s per group in CoCA, 0.10s per group in CoSOD3k, and 0.20s per group in CoSal2015. 
We also evaluate the inference time and speed on the benchmark proposed by \cite{MCCL}\footnote{https://github.com/ZhengPeng7/CoSOD\_fps\_collection. The original benchmark is used on an A100, while we evaluate all models on a 3080Ti to get the results in Table~\ref{tab:stat}.}, and compare them with recent models. Following the original papers, we set the input shape of DCFM~\cite{DCFM} and our models to $224\times224$, and the input shape of GCoNet+~\cite{GCoNet+}/MCCL~\cite{MCCL}/DMT~\cite{DMT} to $256\times256$. We also utilize fvcore to calculate the FLOPs of these models\footnote{The batch size is set to 1 for all models when checking FLOPs.}, and count the number of their parameters. The results are shown in Table~\ref{tab:stat}. Though our model is heavier than previous ones in terms of parameters, its computational complexity does not show a significant increase compared to DCFM~\cite{DCFM} and GCoNet+~\cite{GCoNet+}. Although MCCL~\cite{MCCL} has a very lightweight structure, its inference speed is slower than our model. In summary, our model outperforms previous models in most evaluation metrics while maintaining competitive inference efficiency.

Table \ref{tab:params-full} illustrates the distribution of parameters and computations across the modules of our model. It can be observed that the encoder occupies a substantial portion of the parameters, while the calculation is concentrated in the encoding and decoding stages.

\begin{table}[h]
  \caption{Detailed statistics of the full model.}
  \label{tab:params-full}
  \begin{tabular}{c|c|c}
    \toprule
    Module & \#params & FLOPs\\
    \midrule
    Encoder & 86.2M & 17.7G\\
    Consensus Extraction & 15.8M & 0.2G\\
    Consensus Dispersion & 42.9M & 6.1G\\
    Decoder & 11.8M & 10.7G\\
    \midrule
    Total & 156.7M & 34.7G\\
  \bottomrule
\end{tabular}
\end{table}


\section{Additional Experimental Results}
The paper has presented the ablation studies on the proposed consensus extraction and dispersion modules. In this section we examine the design of other components in the pipeline. As in the paper, CoCA~\cite{GICD} is used as test set, since it contains the most challenging examples and thus can clearly showcase the impact of different designs. Due to time constraints, each ablation experiment is conducted only once. Although there may be some randomness, the trend of the results is quite evident in most cases.

\subsection{Encoder Pre-training Scheme}
We first focus on the encoder, and analyze the impact of CLIP pre-training. As shown in Table~\ref{tab:encoder}, we compare the default pre-training and fine-tuning scheme with three other settings: training from scratch, using the pre-trained weights without fine-tuning, using the pre-trained weights and fine-tuning it with the learning rate setting to the same value as other modules. It can be clearly seen that the default setting achieves the best performance, which shows the importance of both the priors from large-scale pre-training and the task-specific information from the downstream fine-tuning.

\begin{table}[h]
  \caption{Impacts of encoder initialization and training scheme.}
  \label{tab:encoder}
  \begin{center}
\renewcommand{\arraystretch}{1.0}
\renewcommand{\tabcolsep}{1.14mm}
\begin{tabular}{r|ccc}
\hline
Encoder & $E_{\xi}^\mathrm{max} \uparrow$ & $S_\alpha \uparrow$ & $F_\beta^\mathrm{max} \uparrow$ \\
\hline
trained from scratch & 0.699 & 0.593 & 0.387 \\
pre-trained \& frozen & 0.759 & 0.645 & 0.527 \\    
pre-trained \& fine-tuned ($0.1\times$ lr) & {\bfseries0.804} & {\bfseries0.707} & {\bfseries0.598} \\
pre-trained \& fine-tuned & 0.742 & 0.646 & 0.478 \\

\end{tabular}
\end{center}
\end{table}

\subsection{Number of Layers}

Next, we carry out the ablation experiment on the number of ILIAT/ITD layers. Note that the number of GSAT layers should always be the number of ILIAT layers plus one (when \#ILIAT>0), so it is not mentioned explicitly here.

The results in Table~\ref{tab:abl-num} indicate that the \emph{existence} of these two modules brings about the most significant performance improvement, while the impact of changing their \emph{number} is relatively small. Concretely speaking, adding ITD layers has a greater benefit than adding GSAT and ILIAT layers. It reveals that the naive model requires more improvements in the dispersion stage rather than the extraction stage, which is consistent with the findings of the ablation experiments in the paper. Moreover, adding more layers to the default setting (\#ILIAT=2, \#ITD=4) will not further improve the performance. This may be due to overfitting caused by the over-complex consensus learning.

\begin{table}[h]
  \caption{Impacts of the number of layers.}
  \label{tab:abl-num}
  \begin{center}
\renewcommand{\arraystretch}{1.0}
\renewcommand{\tabcolsep}{1.14mm}
\begin{tabular}{c|c|ccc|c}
\hline
\#ILIAT & \#ITD & $E_{\xi}^\mathrm{max} \uparrow$ & $S_\alpha \uparrow$ & $F_\beta^\mathrm{max} \uparrow$ & \#params\\
\hline
0 & 0 & 0.804 & 0.707 & 0.598 & 99.6M\\
2 & 2 & 0.828 & 0.734 & 0.623 & 140.1M\\
2 & 4 & {\bfseries0.831} & {\bfseries0.741} & {\bfseries0.631} & 156.7M\\
4 & 2 & 0.830 & 0.736 & 0.623 & 152.8M\\
4 & 4 &  0.825 & 0.727 & 0.615 & 169.3M\\

\end{tabular}
\end{center}
\end{table}

\subsection{Loss Functions}

Table~\ref{tab:abl-loss} shows the ablation study on the loss functions. Two loss functions are employed in our model, namely the Binary Cross Entropy (BCE) loss and the Intersection over Union (IoU) loss. 
As mentioned in \cite{GCoNet+}, IoU loss supervises the model on the region level, while BCE loss helps the model focus
on details. Experiments also show that combining these two losses leads to better performance than using either one independently.

\begin{table}[h]
  \caption{Impacts of the loss functions.}
  \label{tab:abl-loss}
  \begin{center}
\renewcommand{\arraystretch}{1.0}
\renewcommand{\tabcolsep}{1.14mm}
\begin{tabular}{c|c|ccc}
\hline
IoU loss & BCE loss & $E_{\xi}^\mathrm{max} \uparrow$ & $S_\alpha \uparrow$ & $F_\beta^\mathrm{max} \uparrow$ \\
\hline
\checkmark & & 0.820 & 0.724 & 0.626 \\
 & \checkmark & 0.822 & 0.728 & 0.613 \\
\checkmark & \checkmark & {\bfseries0.831} & {\bfseries0.741} & {\bfseries0.631}  \\

\end{tabular}
\end{center}
\end{table}

\subsection{Iterative Refinement}
As suggested by the reviewer, we carry out an experiment on the effectiveness of the iterative refinement (in the consensus extraction module). The iterative refinement uses multiple GSAT-ILIAT process. It means that the extracted consensus from one GSAT layer is further refined by the following ILIAT layer and GSAT layer, utilizing local details obtained from the hierarchical structure. Table~\ref{tab:modules} has already proven the effectiveness of GSAT and ILIAT separately. Here we compare the default full model (with 3 GSAT layers and 2 ILIAT layers) against a model with less iterations (i.e. 2 GSAT layers and 1 ILIAT layer). The results are shown in Table~\ref{tab:abl-iter}. It can be concluded that introducing the iterative process in the consensus extraction stage can better refine the consensus representation with local details, thus enhance the overall performance.

\begin{table}[h]
  \caption{Impacts of iterative refinement.}
  \label{tab:abl-iter}
  \begin{center}
\renewcommand{\arraystretch}{1.0}
\renewcommand{\tabcolsep}{1.14mm}
\begin{tabular}{c|c|ccc}
\hline
\#GSAT & \#ILIAT & $E_{\xi}^\mathrm{max} \uparrow$ & $S_\alpha \uparrow$ & $F_\beta^\mathrm{max} \uparrow$\\
\hline
2 & 1 & 0.828 & 0.730 & 0.625 \\
3 (default) & 2 (default) & {\bfseries0.831} & {\bfseries0.741} & {\bfseries0.631} \\

\end{tabular}
\end{center}
\end{table}

\end{document}